\documentclass[sigconf ]{acmart}
\usepackage{subfigure}
\usepackage{placeins} 

\AtBeginDocument{%
  \providecommand\BibTeX{{%
    \normalfont B\kern-0.5em{\scshape i\kern-0.25em b}\kern-0.8em\TeX}}}

\setcopyright{acmlicensed}
\copyrightyear{2024}
\acmYear{2024}
\acmDOI{XXXXXXX.XXXXXXX}

\acmConference[SIGKDD]{Make sure to enter the correct
  conference title from your rights confirmation emai}{June 03--05,
  2018}{Woodstock, NY}
\acmISBN{978-1-4503-XXXX-X/18/06}

\begin{document}

\title{Scaling Laws for Discriminative Classification in Large Language Models}


\author{Dean Wyatte}
\affiliation{%
  \institution{Cash App}
  \city{Boulder}
  \state{Colorado}
  \country{USA}
}
\email{dwyatte@block.xyz}

\author{Fatemeh Tahmasbi}
\affiliation{%
  \institution{Cash App}
  \city{Woodbridge}
  \state{Virginia}
  \country{USA}
}
\email{fatemeh@block.xyz}

\author{Ming Li}
\affiliation{%
  \institution{Cash App}
  \city{Bellevue}
  \state{Washington}
  \country{USA}
}
\email{victorl@block.xyz}

\author{Thomas Markovich}
\authornote{Corresponding Author}
\authornotemark[1]
\affiliation{%
  \institution{Cash App}
  \city{Cambridge}
  \state{Massachusetts}
  \country{USA}}
\email{tmarkovich@block.xyz}

\renewcommand{\shortauthors}{Wyatte, et al.}

\begin{abstract}
Modern large language models (LLMs) represent a paradigm shift in what can plausibly be expected of machine learning models. The fact that LLMs can effectively generate sensible answers to a diverse range of queries suggests that they would be useful in customer support applications. While powerful, LLMs have been observed to be prone to hallucination which unfortunately makes their near term use in customer support applications challenging. To address this issue we present a system that allows us to use an LLM to augment our customer support advocates by re-framing the language modeling task as a discriminative classification task. In this framing, we seek to present the top-K best template responses for a customer support advocate to use when responding to a customer. We present the result of both offline and online experiments where we observed offline gains and statistically significant online lifts for our experimental system. Along the way, we present observed scaling curves for validation loss and top-K accuracy, resulted from model parameter ablation studies. We close by discussing the space of trade-offs with respect to model size, latency, and accuracy as well as and suggesting future applications to explore.
\end{abstract}



\maketitle

\section{Introduction}
Modern Large Language Models (LLMs) such as those in the GPT~\citep{radford2019language, brown2020language, achiam2023gpt}, LLaMA~\citep{touvron2023llama, touvron2023llama2}, PaLM~\citep{chowdhery2023palm}, and Pythia~\citep{biderman2023pythia} families are decoder-only models employing the transformer architectures, and are trained to perform next token prediction through an autoregressive objective. These models work by using attention with a causal mask  to learn a representation of a sequence of tokens that can then be used to predict the next best token. Because these models are decoder-only, they are significantly simpler and may be more sample efficient ~\citep{tay2022unifying,wang2022language} meaning that much larger models can be trained on drastically more data. It has been shown that LLM training exhibits strong scaling laws that relate the size of the model and the available data to the overall model quality on a variety of downstream tasks~\citep{achiam2023gpt}. The existence of these scaling curves has shown that our current LLMs are not yet at risk of overfitting, causing us to conclude that we can improve both by just increasing the model size and amount of training data~\citep{hoffmann2022training, kaplan2020scaling}. To date, the vast majority of LLM work has been focused on open-domain tasks, but for industrial settings, closed domain tasks are more valuable in the near term. These are tasks for which the action and topic spaces are drastically reduced. Customer service applications are a common application for language modeling, because they typically involve extracting nuanced meaning from multi-turn text based conversations and are a common part of many corporate product offerings. Indeed, the customer service application market is projected to be as large as 58 billion dollars ~\citep{acumen2023csmarket} with multiple existing offerings such as Google's DialogFlow and Amazon's Lex.

Given the generative nature of LLMs, an autonomous customer support system is a natural application of an LLM. While LLMs seem uncontested in many applications, they are not without their problems. In generative settings, these models have been shown to hallucinate answers, and are susceptible to data exfiltration attacks. In customer support use cases, both of these failure modes preclude adoption. Beyond these concerns, LLMs can also have the potential to reply to focused queries from customers with off-topic or innane responses which would ultimately lead to degraded customer experience. Additionally, LLMs are also extremely computationally demanding which makes their deployment in high throughput product applications expensive or untenable. Solving these problems is the focus of intense research across the community.

Rather than waiting for these problems to be solved, we ask the question \textbf{``How can we use LLMs to solve intermediate, industrially relevant, problems?''} We find this answer in using LLMs to perform next-reply classification given a set of pre-written replies for our chat support system as a way to augment and speed up our customer support representatives. We will refer to this problem as template classification, which we frame as a supervised machine learning task for a decoder-only LLM architecture that we fine-tune into a discriminative classifier. Framing the problem in this way allows us to enjoy the well documented benefits of modern language models while avoiding the well documented risks associated with model hallucination and data leakage. With this framing in hand, we construct a two phase training pipeline that first domain-adapts the LLM and then fine-tunes it into a discriminative classifier.  As this is an industrial system, we consider the correspondence between offline machine learning metrics and online product metrics which we assess through rigorous online experimentation. We additionally explore the trade-off between accuracy, which generally correlates with model accuracy, and run-time performance through a model parameter ablation. In summary, this paper's core contributions are:

\begin{enumerate}
  \item We detail to our knowledge the first industrial implementation of fine-tuning LLMs into discriminative classifiers.
  \item We detail to our knowledge the first scaling law on closed domain adaptation.
  \item We detail the relationship between language modeling performance and amenability to discriminative fine-tuning for a classification task with hundreds or thousands of classes. 
  \item We discuss practical considerations associated with putting these models into production.
\end{enumerate}

While this system focuses on augmenting our customer advocates with LLMs, we view this as the first in a series of steps towards deploying a reliable, autonomous, customer support agent. The remainder of this paper is organized as follows. In Section 2, we review related work with a particular focus on scaling laws and using LLMs for classification. In Section 3, we review our system. We detail our Methods in Section 4. We present our offline and online experimental results in Section 5, and carefully discuss the practical considerations associated with implementing and deploying these models within our system. And finally, we present our conclusions and next steps in Section 6.

\section{Related Work}
In this section we situate our work within the large language model literature, which has rapidly evolved over the last few years.

\textbf{Language Models and Customer Support.} Language modeling has frequently been used to construct chat interfaces that assist customers in handling common queries in a faster and cheaper way than human agents can provide, and in 2023 Gartner found that chatbots were used to resolve 58\% of all billing disputes~\citep{gartner2023csaas}. These chat systems are often built on a rigid set of rules and business logic, and language modeling is used for intent classification to navigate the tree. Because encoder-decoder models such as BERT provide a trained encoder which can be used for downstream classification, many of these intent classification systems were built with BERT as a backbone. In 2021 CS-BERT, BERTa$\backslash$'u, and AVA were released, which are a BERT variants that were finetuned on customer support data, and provided significant improvements across a variety of customer support benchmarks~\citep{wang2021cs, finardi2021berta, yu2021ava}. Recently, generative language models have been explored within a customer support use case~\citep{pandya2023automating, shahin2024harnessing}, but are not without their very public failures due to hallucination of company policies~\citep{Proctor_2024} or PII leakage~\citep{Burgess_2023}. To the best of our knowledge, BERT-based models remain the state of the art for intent classification based systems.

\textbf{Large Language Models.} Almost all modern language models use a transformer architecture~\citep{vaswani2017attention}. At the heart of this architecture is the transformer block which is comprised of multi-head self attention~\citep{vaswani2017attention}, layer normalization units~\citep{ba2016layer}, a dense fully connected unit, and residual connections. Modern transformers are constructed by stacking many of these blocks, and in the instance of text, the transformer ingests input tokens and outputs predicted tokens based on some training objective. Most modern LLMs follow a causal decoder-only architecture, such that the LLM has no way to learn different representations for input and target sequences. Instead, through the use of causal masking, the decoder-only transformer predicts the next token conditioned on the previously observed tokens. Popular examples of this model structure include GPT~\citep{radford2019language, brown2020language, achiam2023gpt}, LLaMA~\citep{touvron2023llama, touvron2023llama2}, PaLM~\citep{chowdhery2023palm}, and Pythia~\citep{biderman2023pythia}.

\textbf{Model Scaling Behavior.} Scaling behaviors have long been observed in the training of language models. The authors of \citet{kaplan2020scaling} were some of the first to empirically observe power-law scaling in dataset volume and model size from their training of GPT2~\citep{kaplan2020scaling}. They additionally observed no instances of overfitting for 1 billion parameters and tokens. In the development of Chinchilla, the authors provided an empirical analytical formula that recommended the optimal model size and data volume for a fixed compute budget~\citep{hoffmann2022training}. There has also been work focused on understanding the scaling behavior of language model fine-tuning. It has previously been observed that in domains with a fixed number of tokens, the optimal strategy is to take a pre-trained model and fine-tune it for the domain at hand~\citep{hernandez2021scaling, gururangan2020don}. In \citet{hernandez2021scaling}, a relationship was derived to predict the loss reduction associated with fine-tuning when compared with training a transformer of the same size from scratch on the same data.

\textbf{Domain Adaptation of Language Models.} There is a significant body of literature that is focused on applying language models to narrow domains through various domain adaptation schemes. An early version of this idea was presented as domain adaptive pre-training (DAPT), which found that a second phase of pre-training on domain-specific datasets led to significant performance gains~\citep{gururangan2020don}. Another early example is that of TOD-BERT, which performed domain adaptation through both dataset curation and loss function augmentation ~\citep{wu2020tod}. The dataset was curated from nine different human-human interaction or multi-turn tasks, and the pre-training used a combination of masked language modeling and contrastive loss that exploited the dialog-specific nature of the pre-training datasets. Using both of these strategies, they were able to realize significant performance gains. It was shown in CS-BERT that in-domain pre-training on millions of customer support messages provided significant benefits on downstream customer support applications~\citep{wang2021cs}. Similar results have been observed in the case of modern decoder-only architectures. In ChipNeMo the authors domain adapted a LLM for use in a chip design which is a highly technical domain~\citep{liu2023chipnemo}. Through domain adaptation, they found that they needed a 5-fold smaller model to achieve same level of performance across a range of downstream tasks~\citep{liu2023chipnemo}. Similar, models like CodeLLaMA~\citep{roziere2023code} and StarCoder~\citep{li2023starcoder} have shown impressive accuracy on a range of programming tasks through a second round or pre-training for domain adaptation. Looking beyond simple domain adaptation, the authors of \citet{cheng2023adapting} developed a language model fine-tuning process inspired by reading comprehension tests. These reading comprehension tests enable the authors to generate domain-specific 7B parameter models that have equivalent quality to 50B parameter models that have been domain adapted through two-phase pre-training. A portion of our work can be understood in this line of investigation because we perform a domain adaptation step.

\textbf{Discriminative Fine-tuning.} There are relatively few example of using decoder transformer architectures for classification tasks through discriminative fine-tuning. Early work with the GPT architecture investigated the effect of language model pre-training on downstream fine-tuning tasks such as text classification, entailment, semantic similarity, and multiple choice question answering ~\citep{radford2018improving}. Subsequent work with the GPT architecture however focused on zero- and few-shot language generation in lieu of fine-tuning. Other similar fine tuning methods could include both Parameter Efficient Fine Tuning (PEFT)~\citep{xu2023parameter} and Supervised Fine Tuning (SFT)~\citep{ouyang2022training}. PEFT refers to a family of methods, such as LoRA~\citep{hu2021lora}, that seek to fine tune a much smaller set of parameters than exist in the original model. SFT seeks to align the original embeddings of the LLM for the target task through a supervised process that maps the LLM output to that of the target. For InstructGPT, this mapping consists of aligning input training text with multiple sequential output tokens. Our method can be understood as a SFT method with an output space that is of size $N_{classes}$. Perhaps the most relevant work is the recent LS-LLaMA approach which systematically studied replacing the final transformer decoder with a classification layer and removing causal masks. In their experiments, they find reliable improvements upon BERT baselines while also outperforming LLMs an order of magnitude larger than LS-LLaMA. Their work however is limited to the LLaMA family of models and they do not quantify the effect of LLM size on discriminative fine-tuning ability.

\section{Preliminaries}

\begin{figure}[h]
	\centering
	\includegraphics[width=0.9\columnwidth]{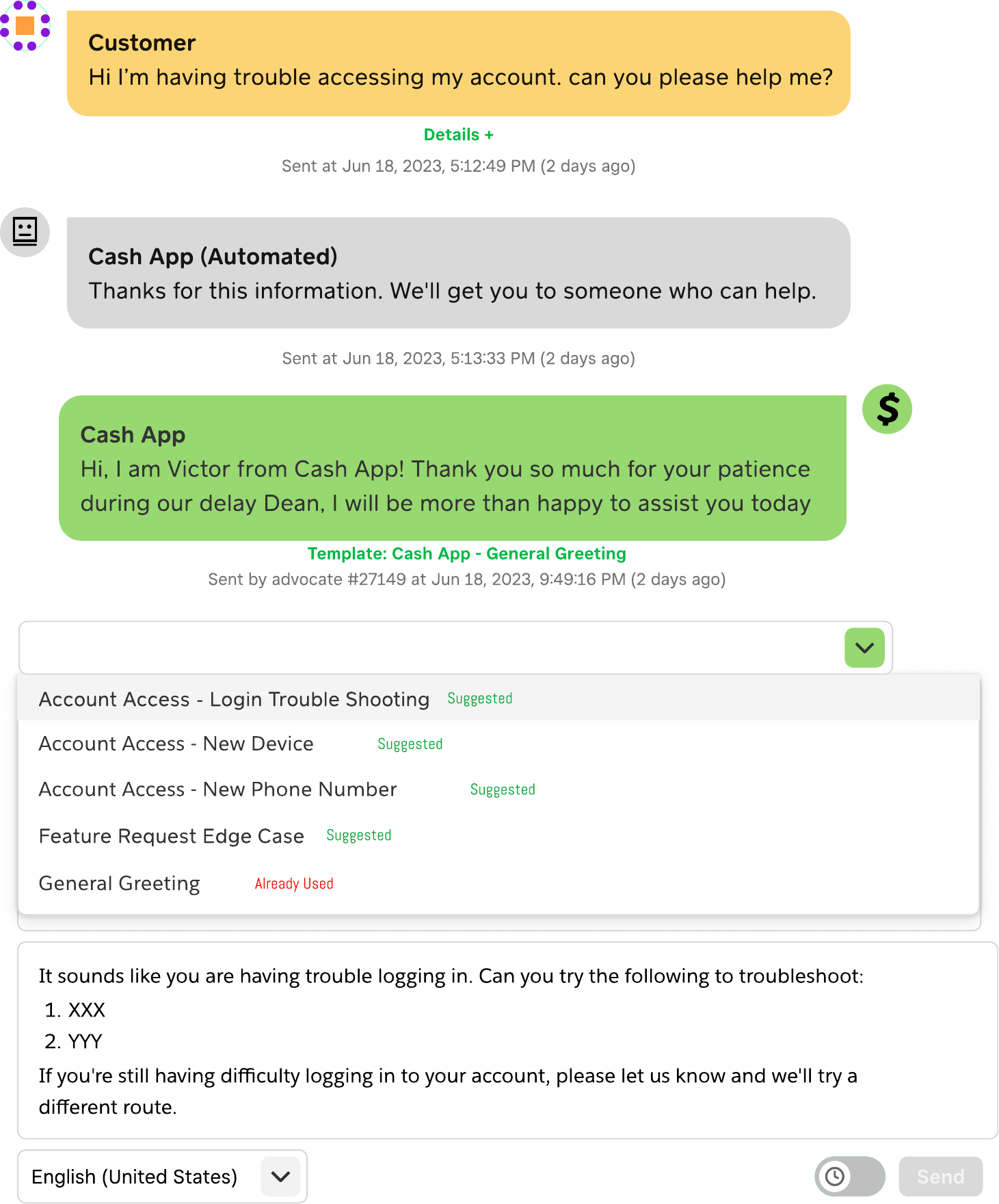}
	\caption{Example customer support case with most relevant template responses.}
	\label{fig:quicktext_example}
\end{figure}

Cash App customer support advocates respond to customer queries in real time using a combination of freehand and template text that they further tailor to a specific customer's query. The templates are used for common customer support actions including greeting the customer, requesting more information about a particular query (e.g., details associated with a financial transaction), troubleshooting steps, and providing status updates about an in-progress support case. Figure \ref{fig:quicktext_example} illustrates typical use of these templates for troubleshooting.

Advocates are trained to choose appropriate templates during a support case, which presents us with an opportunity to automatically suggest the most appropriate ones using a machine learning model that considers the current support case context. To augment the advocates, we have designed a machine learning system that selects the top-k most likely templates that would address the customer's message. In this setting, it is only the advocate who sees the top-k signals. Advocates can then further tailor the suggested templates before responding, dismiss them to manually search for a more appropriate template, or respond completely in freehand text. We describe our modeling approaches to this problem in the following section followed by a broader study of this online system in Section \ref{sec:case_study}.

\section{Methods}
\begin{figure*}[h]
	\centering
	\includegraphics[width=500pt]{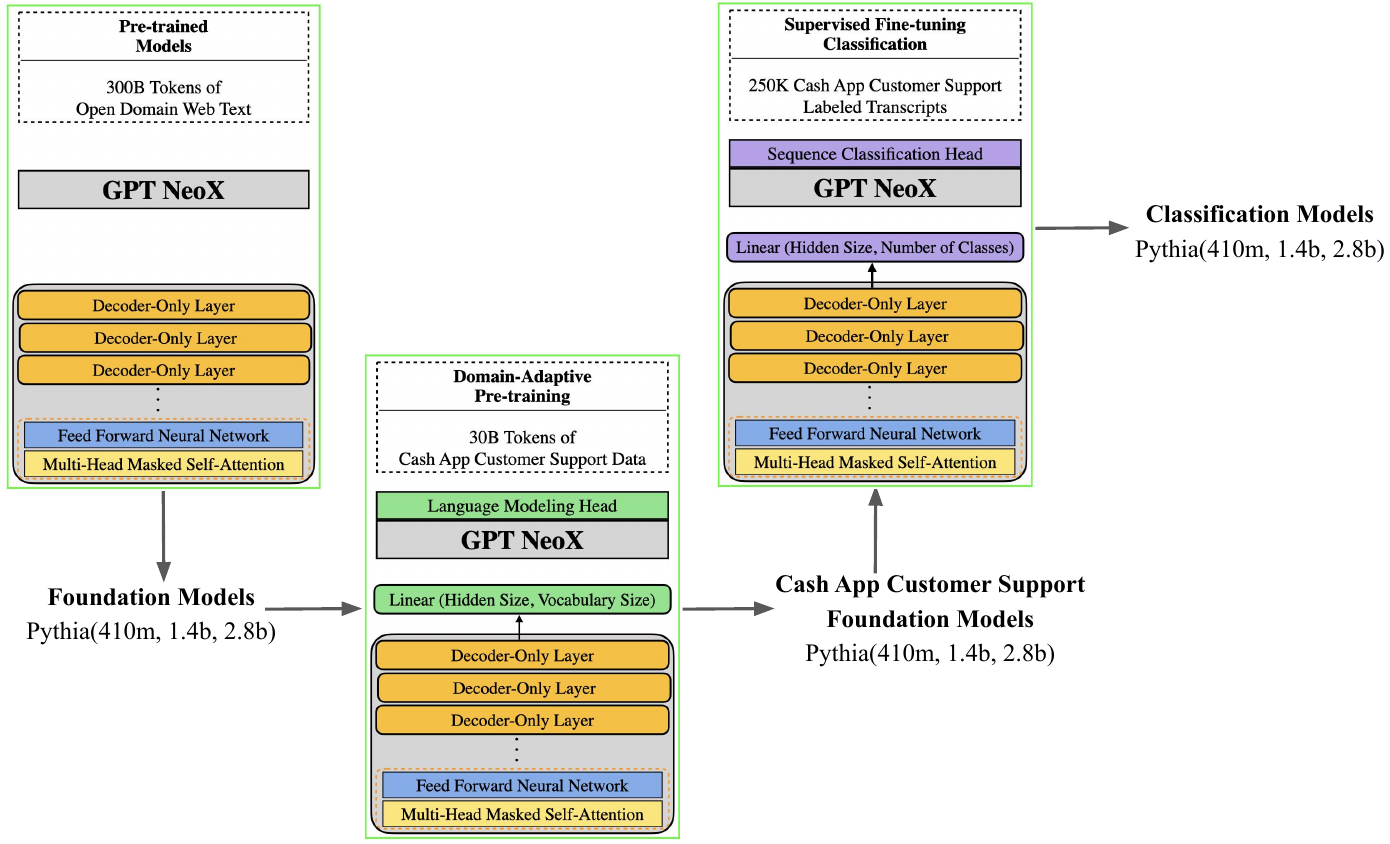}
	\caption{Training pipeline for domain adaptation and discriminative fine-tuning.}
	\label{fig:pipeline}
\end{figure*}

The overall training pipeline is depicted in Figure \ref{fig:pipeline} and consists of two steps: 1) domain adaptation with an autoregressive language modeling objective and 2) discriminative fine-tuning with discrete labels. In the domain adaptation step, we start with a pre-trained LLM and continue pre-training it on data from our target domain consisting of Cash App customer support transcripts. We use the Pythia suite of LLMs ~\citep{biderman2023pythia} which build on the GPTNeoX architecture, and are closely related to the original GPT-3 paper, ~\citep{gao2020pile} with parameter counts from 70M to 12B pre-trained on approximately 300B tokens of domain general web data from The Pile ~\citep{gao2020pile,biderman2022datasheet}. This choice of pre-trained LLM allows us to leverage the effectiveness of transfer learning ~\citep{hernandez2021scaling} while still studying their scaling laws using a range of parameter counts within a closed domain. For this study, we focus on the 410 million (410m), 1.4 billion (1.4b), and 2.8 billion (2.8b) parameter variants to provide a comparison to BERT-large ~\citep{devlin2018bert} (340M parameters), which is commonly used in discriminative fine-tuning while documenting scaling effects for domain adaptation with higher parameter counts.

In the discriminative fine-tuning step, we initialize a new linear layer of size $N_{hidden} \times N_{classes}$ on top of the final transformer block of one of our domain adapted models. We fine-tune the model end-to-end on a smaller dataset of Cash App customer support transcripts labeled with support advocates template response selections to produce our final classifier.

\subsection{Customer Support Dataset} 

Our customer support dataset consists of tens of billions of tokens of customer support transcripts collected using an in-app messaging interface over the course of Cash App's operational history. These transcripts have been processed to remove PII using an industry standard redaction pipeline. The cases are initiated by a message from a customer and contain a combination of automated system responses and human responses as the support advocate works to resolve the customer's query. Examples of how the dataset is formatted in each phase of our training pipeline are shown in the Appendix.

Even though we redact our data, PII leakage is a common risk with LLMs in deployed applications but it is not a risk for our system as designed for two reasons. The first is that we are using our LLM as a classifier to select the correct template to respond to a customer with; and the second is that all responses are routed through a customer service advocate. 

\subsection{Domain Adaptive Pre-training}

We annotate each message in the aforementioned customer support transcripts with a "<CUSTOMER>", "<SYSTEM>", or "<ADVOCATE>" prefix to indicate the participant and join the annotated messages together with newlines. For tokenization, we employ the same BPE tokenizer used in the pre-trained Pythia model opposed to any domain-specific tokenization such as in ~\citep{liu2023chipnemo}. Documents shorter than 2048 tokens are packed together into full length sequences to increase throughput. Shorter documents within the sequence are separated with a special end-of-text token to give the model indication that these documents are unrelated.

During domain adaptation, we generally follow the original pre-training configuration for the model sizes reported in ~\citep{biderman2023pythia} including the use of the AdamW optimizer with $\beta_{1}$ and $\beta_{2}$ values of 0.9 and
0.95, and weight decay of 0.01, learning rate based on the model size, and batch size of 2 million tokens. We employ a linear warmup for 1\% of our total training steps followed by cosine decay to zero. Our complete hyperparameters are shown in Table~\ref{tab:da_hyperparameters}. Additionally, we use the Zero Redundancy Optimizer (ZeRO) ~\citep{rajbhandari2020zero} to efficiently scale training to multiple GPUs.

\begin{table}[ht]
	\centering
	\begin{tabular}{lcc} 
		\hline
		{\textbf{Model}}& 	{\textbf{Configuration Key}} & \textbf{Value} \\
		\hline
		Pythia 410m, 1.4b, 2.8b&&\\
		&fp16.enabled & True \\
		&lr-decay-style & cosine \\
		&max-position-embeddings & 2048 \\
		&optimizer.params.betas &  [0.9, 0.95] \\
		&optimizer.type & AdamW \\
		&warmup & 0.01 \\
		&weight-decay & 0.01 \\
		&max-steps & 14500 \\
		&eval-steps & 0.1 \\
		&save-steps & 0.1 \\
		\hline
		Pythia 410m& learning rate & 3e-4 \\
		Pythia 1.4b& learning rate & 2e-4 \\
		Pythia 2.8b& learning rate& 1.6e-4 \\
		\hline
	\end{tabular}
	\caption{Configuration details for Pythia models domain adaptation. Common configurations are listed above the middle line, with model-specific configurations below.}
	\label{tab:da_hyperparameters}
\end{table}

We pre-train for up to 14500 steps of 2M tokens, reserving 1B tokens for evaluation although due to computational constraints and time considerations, we interrupted the larger models late in training. We save checkpoints every 1450 steps (2.9B tokens) which we use for discriminative fine-tuning.

\subsection{Discriminative Fine-tuning}

In this step we fine-tune the domain-adapted Pythia models from the previous step for a sequence classification task. We select 250K random messages on which a customer support advocate chose to use a template reply and consider that as our label. We use a 50\% train/test split at the support case level to prevent leakage if selecting multiple messages from the same support case. The label set comprises 640 unique template responses.

We consider the support case up to the labeled message as the context. We are interested in comparing our domain adapted models with BERT-large which has a maximum context of 512 tokens, so we use a rolling window with an earliest-first truncation strategy to select the most recent whole messages up to the maximum of 512 tokens. We omit the "<CUSTOMER>", "<SYSTEM>", and "<ADVOCATE>" annotations in this case to give more room for support case context based on previous unpublished work that shows these annotations provide minimal information in this task.

We initialize a new linear layer of 640 classes and select the right-most token of each sequence in the batch to pool as input as in ~\citep{li2023label} and implemented in the transformers library \cite{wolf-etal-2020-transformers}. We fine-tune for one epoch as we have observed additional fine-tuning quickly overfits and evaluate the result with Top-1, Top-3, and Top-5 accuracy.

Fine-tuning hyperparameters are shown in Table~\ref{tab:ft_hyperparameters}
\begin{table}[ht]
	\centering
	\begin{tabular}{lcc} 
		\hline
		{\textbf{Model}}& 	{\textbf{Configuration Key}} & \textbf{Value} \\
		\hline
		BERT-large, && \\ Pythia 410m, 1.4b, 2.8b&&\\
		&fp16.enabled & True \\
		&lr-decay-style & linear \\
		&max-position-embeddings & 512 \\
		&optimizer.params.betas &  [0.9, 0.99] \\
		&optimizer.type & AdamW \\
		&warmup & 0.1 \\
		&weight-decay & 0.0 \\
		&learning rate & 1e-5 \\
		&batch size & 128 \\
		\hline
		BERT-large& num-train-epochs & 10 \\
		Pythia 410m, 1.4b, 2.8b& num-train-epochs & 1 \\
		\hline
	\end{tabular}
	\caption{Configuration details for discriminative fine-tuning. Common configurations are listed above the middle line, with model-specific configurations below.}
	\label{tab:ft_hyperparameters}
\end{table}

\subsection{Model Updates}
Our model training pipeline is composed of two phases -- domain adaptation and discriminative fine tuning. The domain adaptation phase requires billions of tokens and can take multiple days to train while discriminative fine tuning requires orders of magnitude fewer tokens and only takes hours. Because domain adaptation is trained using the standard causal autoregressive objective, we find that we are able to reuse domain adapted models across different discriminative fine tuning jobs. We anticipate performing further domain adaptation after 10 billion additional tokens have been acquired, but otherwise update models via discriminative fine tuning to accommodate changes to template usage.

\subsection{Baseline Models}

As a baseline, we fine-tune BERT-large as well as customer support domain-adapted variant using the approach described in ~\citep{gururangan2020don} that is used in several of our other online systems. For details of our domain adaptation using BERT's masked language modeling objective, we refer readers to ~\citep{cashAIBlogSupport}. Given BERT-large generally takes several epochs to converge ~\citep{devlin2018bert}, we fine-tune for a maximum of 10 epochs and report the epoch with the highest test set performance which which in practice, was epoch 9.

\section{Experiments}
\subsection{Offline Training and Evaluation}
\begin{figure*}[ht]
	\centering     
	\subfigure[Domain Adaptation FLOPs vs Classification Loss]{\label{fig:a}\includegraphics[width=57mm]{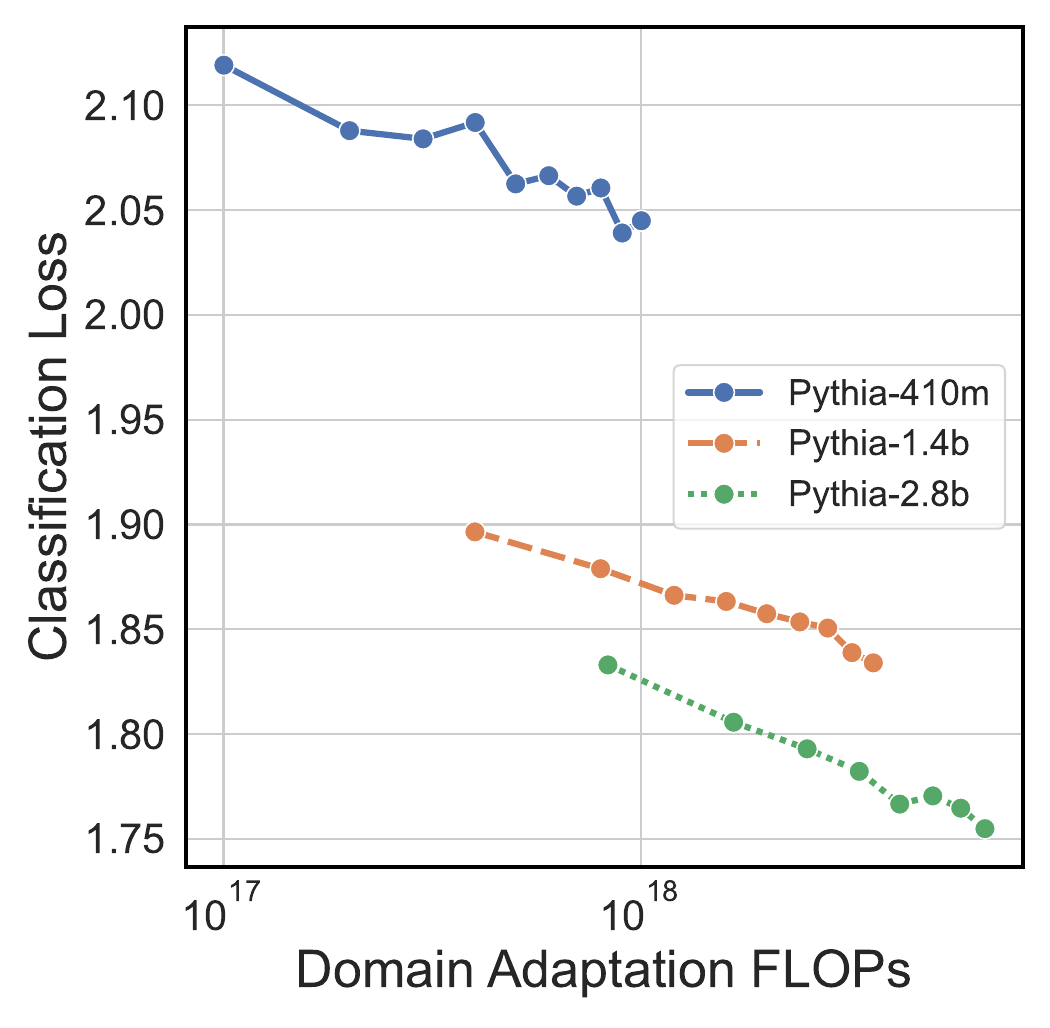}}
	\subfigure[Domain Adaptation Tokens vs Classification Loss]{\label{fig:b}\includegraphics[width=56mm]{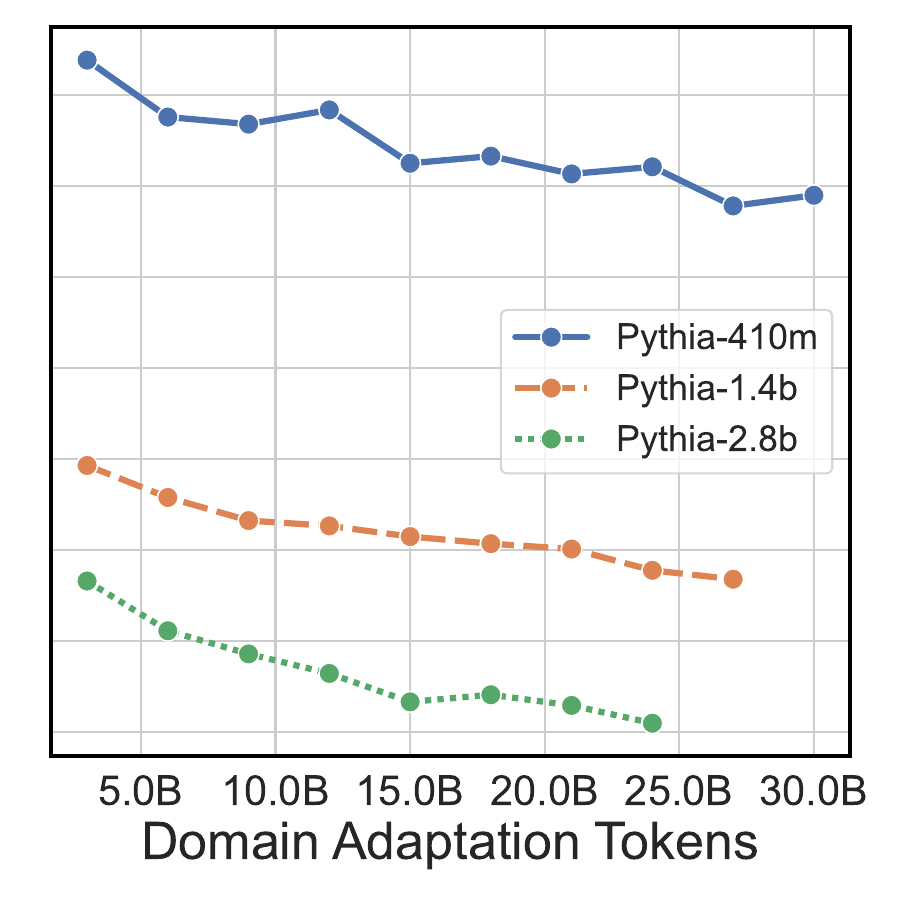}}
	\subfigure[Domain Adaptation Loss vs Classification Loss]{\label{fig:c}\includegraphics[width=56mm]{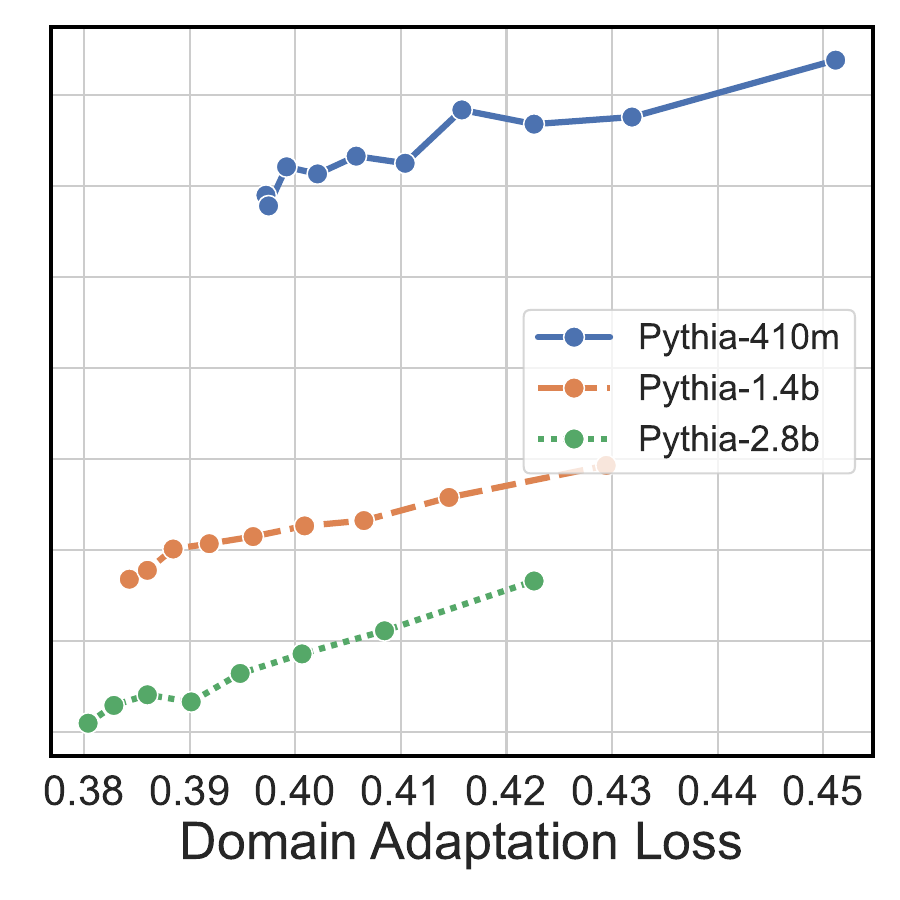}} 
	\caption{Discriminative fine-tuning empirical scaling properties across different model sizes. }
	\label{fig:empricalScaling}
\end{figure*}



Given our two-stage training pipeline, our goal in studying scaling laws is to predict how well a model adapted to a domain using a language modeling objective (stage 1) will perform when discriminatively fine-tuned as a classifier (stage 2). We evaluate this scaling behavior using various metrics that offer insights into performance, efficiency, and behavior across different scales, including

\textbf{FLOPs (Floating Point Operations):} The number of floating point operations executed during the language modeling stage. It provides insights into the computational complexity of the LLM and how it scales with model and dataset size.

\textbf{Language Modeling Loss:} The cross entropy between the model's predicted next token and the actual next token in training data, directly optimized by the language modeling objective. Monitoring the loss ensures that the model is converging toward an optimal language model of the data.

\textbf{Classification Loss:} The cross entropy between the model's predicted class probabilities and the ground truth label, directly optimized by the discriminative fine-tuning objective. It provides a measure of how well the classifier can differentiate between classes.

Figure~\ref{fig:empricalScaling} depicts properties of the scaling laws observed in our experiments. In Figure~\ref{fig:empricalScaling} (a) we plot classification loss as a function of language modeling FLOPs for each of the model sizes we adapted. We observe overlap across the three model sizes tested such that for a given compute budget such as $10^{18}$ FLOPs,  larger models exhibit lower training loss, suggesting their ability to learn complex patterns and structure within the training data. While we do not fit a formal scaling law here, the observation is consistent with compute optimal language modeling ~\citep{kaplan2020scaling, hoffmann2022training} applied to discriminative fine-tuning.

In Figure~\ref{fig:empricalScaling} (b), we plot the classification loss as a function of number of tokens seen during our domain adaptation stage. We observe scaling as a function of number of training tokens with larger models exhibiting lower classification loss for the same number of training tokens. We also observe a linear relationship between supervised classification loss and the number of training tokens across model sizes. These results suggest that even the largest model we trained will still benefit from more data. We find this encouraging because it means we are able to produce better classifiers when discriminatively fine-tuned, despite our practical parameter-limitations~\citep{kaplan2020scaling, hernandez2021scaling}.

We relate language modeling loss to classification loss in Figure~\ref{fig:empricalScaling} (c). We observe a linear relationship between language modeling loss during domain adaptation and classification loss during discriminative fine-tuning across model sizes. This means despite the different training objectives in these two stages of our pipeline, we can easily predict how amenable a particular LLM is to discriminative fine-tuning from its language modeling abilities. One reason for this may be due to our choice of pooling during discriminative fine-tuning using the right-most token of the input sequence. This makes our fine-tuning objective to minimize $p(y_{i} | x_{0}...x_{n-1})$. If we consider class $y_{i}$ to be a special "class token", the fine-tuning objective reduces to the standard language modeling objective assuming all other regular subword tokens $x$ are masked.

In Table~\ref{tab:accuraciesCompTable} we compare the classification accuracy for different model sizes with BERT-large serving as a baseline. The results span a wide accuracy range of more than 10\% between the weakest model (BERT-large) and the strongest model (Domain Adapted Pythia-2.8b), highlighting the impact of scaling data used for language modeling as well as increasing model size. Domain adaptation regularly improves the accuracy of a model by approximately 4-5\%, which is a larger uplift than that which results from increasing the model size at the scales tested here. We observe that the Pythia models outperform BERT in the roughly parameter equivalent case of BERT-large (310m) compared to Pythia 410m with Pythia 1.4b outperforming BERT-large with domain adaptation. This comparison not only underscores the impact of model and dataset size on accuracy, but may also point to the nature of the language modeling objective in determining discriminative fine-tuning performance. BERT-based models are pre-trained with a masked language modeling objective that only covers 15\% of tokens while the causal language modeling objective used in the Pythia models predicts every token in the sequence which may make language modeling more sample efficient ~\citep{tay2022unifying,wang2022language} and allow use of larger datasets. Finally, while we limit the sequence length to 512 tokens during discriminative fine-tuning to ensure a fair comparison between BERT-large and Pythia models, the Pythia models can make use of up to 2048 tokens of context during language modeling to learn longer-range dependencies.




We also quantify the latency of a typical forward pass for the model sizes we consider which is important for deployed use cases that contain a human in the loop like customer support. While transformer latencies are well-described by the model architecture itself including number of parameters ~\cite{korthikanti2023reducing}, we compute them empirically for our observed sequences via load test in our production environment. We optimize each model using TensorRT on an NVIDIA A10 GPU with FP16 support and then conduct a load test for five minutes sampling from a distribution of 10000 input sequences. The load test makes requests of batch size 1 at a given rate to our end-to-end inference service that includes truncation to a maximum length of 512 tokens. Table~\ref{tab:PredictionTime} lists the peak 1-minute average, P99, and max latencies in milliseconds observed during the load test. We observe that Pythia-2.8b quickly saturates a single GPU at higher levels of concurrency leading to increased tail latency. Because our production system contains a human in the loop that is shown the model's predictions, we establish a latency budget as to not negatively impact their workflow. For a latency budget of 100 ms, Pythia-2.8b will need to be scaled to an additional GPU for each additional request per second, which we view as impractical for most industrial use cases. In contrast, Pythia-1.4b on a single GPU can handle 5-10 requests per second and Pythia-410m fails to saturate a single GPU at the request rates considered here. While larger models may thus appear to be less well-suited for use cases with low latency budgets, we see promise in the relatively smooth linear scaling of classification loss with number of training tokens (Figure 3, for example Pythia-1.4b trained over approximately 27B tokens has approximately the same classification loss as Pythia-2.8 trained over approximately 3B tokens). Therefore domains with a large amount of unlabeled data may be able to utilize smaller models trained for longer or even by repeating data~\citep{muennighoff2023scaling}.
\begin{table}[ht]
	\centering
	\begin{tabular}{l c c c}
		\hline
		\textbf{Model} &  \textbf{Top-1} & \textbf{Top-3} & \textbf{Top-5}  \\
		& \textbf{Acc. (\%)} & \textbf{Acc. (\%)} & \textbf{Acc. (\%)} \\
		\hline
		BERT-large&  46.39 & 66.68 & 74.24  \\
		~ \textit{+ Domain Adaptation} & 49.24 &  70.51	& 78.04  \\
		\hline
		Pythia-410m & 45.43 & 66.96 & 75.02 \\
		~ \textit{+ Domain Adaptation} & 51.09  & 73.31 & 80.71  \\
		Pythia-1.4b & 48.64 & 70.87 & 78.89 \\
		~ \textit{+ Domain Adaptation} &  53.75 & 76.59  & 83.95 \\
		Pythia-2.8b & 49.96 & 72.33 & 80.31 \\
		~ \textit{+ Domain Adaptation} &  \textbf{55.12} & \textbf{78.02} & \textbf{85.10}  \\
		\hline
	\end{tabular}
	\caption{Comparison of classification accuracy across different models after discriminative fine-tuning.}
	\label{tab:accuraciesCompTable}
\end{table}

\begin{table}[ht]
    \centering
    \begin{tabular}{l l l l}
    \hline
        \textbf{Requests/Sec} & \textbf{Pythia-410m} & \textbf{Pythia-1.4b} & \textbf{Pythia-2.8b} \\ \hline
        1 & Avg: 14.29 & Avg: 28.80 & Avg: 45.92 \\ 
        ~ & P99: 30.72 & P99: 40.95 & P99: 70.66 \\
        ~ & Max: 32.37 & Max: 41.45 & Max: 72.08 \\ \hline
        2 & Avg: 13.97 & Avg: 31.64 & Avg: 51.42 \\ 
        ~ & P99: 28.90 & P99: 59.91 & P99: 111.43 \\
        ~ & Max: 29.58 & Max: 67.07 & Max: 118.84 \\ \hline
        5 & Avg: 13.66 & Avg: 30.96 & Avg: 52.72 \\ 
        ~ & P99: 26.21 & P99: 60.65 & P99: 121.27 \\ 
        ~ & Max: 33.85 & Max: 66.67 & Max: 149.56 \\ \hline
        10 & Avg: 14.05 & Avg: 33.02 & Avg: 63.89 \\
        ~ & P99: 31.84 & P99: 79.86 & P99: 183.26 \\
        ~ & Max: 59.33 & Max: 116.40 & Max: 222.36 \\ \hline
        20 & Avg: 15.03 & Avg: 40.42 & Avg: 122.38 \\ 
        ~ & P99: 34.55 & P99: 110.96 & P99: 371.90 \\
        ~ & Max: 46.92 & Max: 178.11 & Max: 482.43 \\ \hline
   \end{tabular}
   \caption{Peak 1-minute average, P99, and max latencies across models.}
   \label{tab:PredictionTime}
\end{table}

\subsection{Online Case Study}
\label{sec:case_study}
We now turn to experiments that we have performed in our online system that uses the models described in preceding sections to surface the most appropriate templates to customer support advocates. Whenever a new message is sent during a customer support case, the model considers the most recent conversation context (up to its maximum context length) and returns the top-5 highest probability templates that are part of its training set as candidate classifications. To evaluate the effectiveness of the system, we remove these predictions from a randomly selected 2\% of advocate-support case interactions which allows us to determine how often an advocate would choose one of our predicted templates when they are working on a given support case as well as the effects of the system on other business metrics.

\begin{figure}[h]
  \centering
  \includegraphics[width=0.45\textwidth]{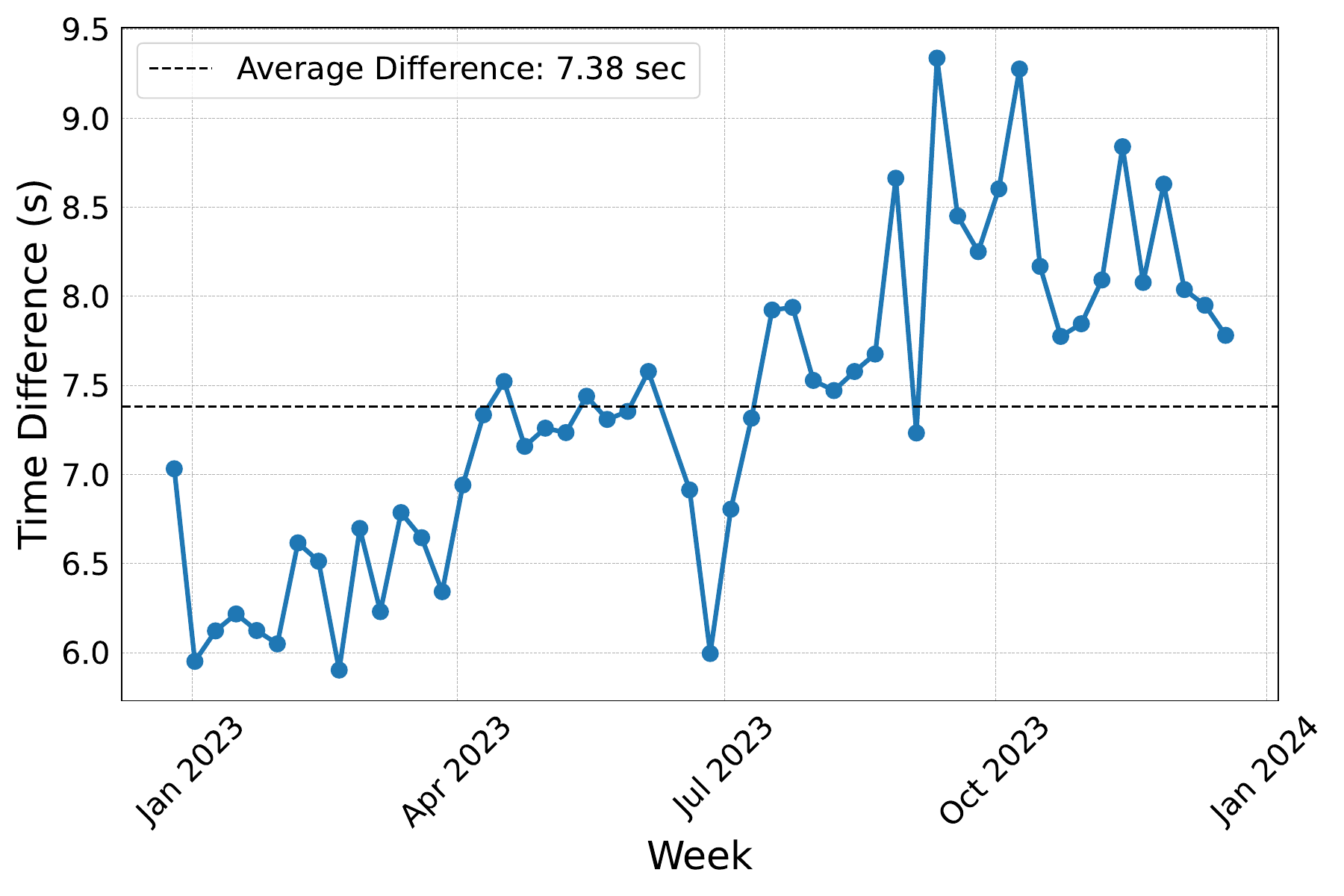}
  \caption{Weekly absolute difference in selection time between holdout and treatment groups.}
  \label{fig:ast_diff}
\end{figure}

Our primary metrics of interest are related to customer support efficiency that do not affect our customer-facing support experience in any noticeable way. A simple first order metric to optimize for is the amount of time it takes an advocate to choose the correct template response based on their training. We find that we have reduced the selection time by 7.38 seconds on average over the lifetime of the system and in general, are able to continually improve on these savings (Figure \ref{fig:ast_diff}). These selection time savings correspond to a 3.56\% total time reduction over the course of an entire support case, a substantial savings when considering the support footprint required for Cash App's tens of millions of active users. During the initial launch of the system, we conducted an internal audit and found no evidence that this time reduction affected customer support standards.

\begin{figure}[h]
  \centering
  \includegraphics[width=0.45\textwidth]{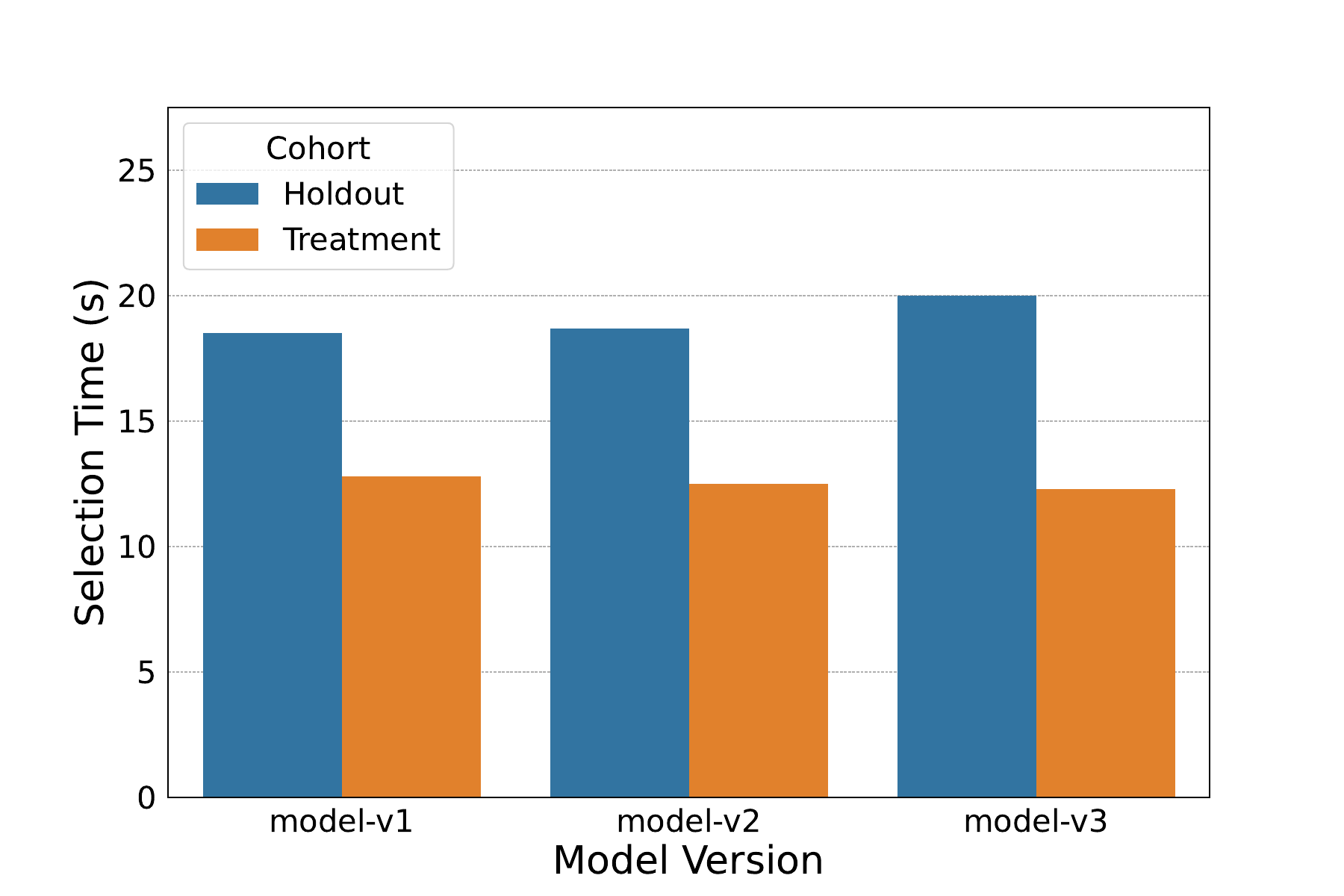}
  \caption{Selection time between holdout and treatment groups over model lifetime}
  \label{fig:model_lifetime}
\end{figure}

New response templates are frequently introduced and existing ones are deprecated, so we need to retrain our model in a discriminative manner. To date, we have retrained our model four time and before releasing the model, we A/B test over a shorter two-week period to determine the effect on our efficiency metrics.\footnote{At time of writing, we have A/B tested our fourth model but have not yet released it.} Over the lifetime of each model, we find that we are consistently able to keep the time it takes to select the correct template with our predictions at approximately 13 seconds, in contrast to 19 seconds without templates (Figure \ref{fig:model_lifetime}). A/B tests indicate that we are able to significantly decrease selection time as a function of retraining (Table \ref{tab:ab_test_model_versions}). Together these results are consistent with the overall increase in selection time savings over the total system lifetime (Figure \ref{fig:ast_diff}).

\begin{table}[h!]
  \centering
   \begin{tabular}{c c c c}
   \hline
   \textbf{A/B Test} &  \multicolumn{2}{c}{\textbf{Selection Time (seconds)}} & \textbf{$p$-value} \\
   			    & \textbf{A} & \textbf{B} \\
   \hline
   model-v1/model-v2 & 13.78 & \textbf{13.33} & 4.76e-10 \\
   model-v2/model-v3 & 12.98 & \textbf{12.52} & 9.09e-10 \\
   model-v3/model-v4 & 11.64 & \textbf{10.91} & 3.67e-27 \\
   \hline
   \end{tabular}
   \caption{A/B tests between existing and retrained models over a two-week period.}
   \label{tab:ab_test_model_versions}
  \end{table}

\begin{figure}[h]
  \centering
  \includegraphics[width=0.45\textwidth]{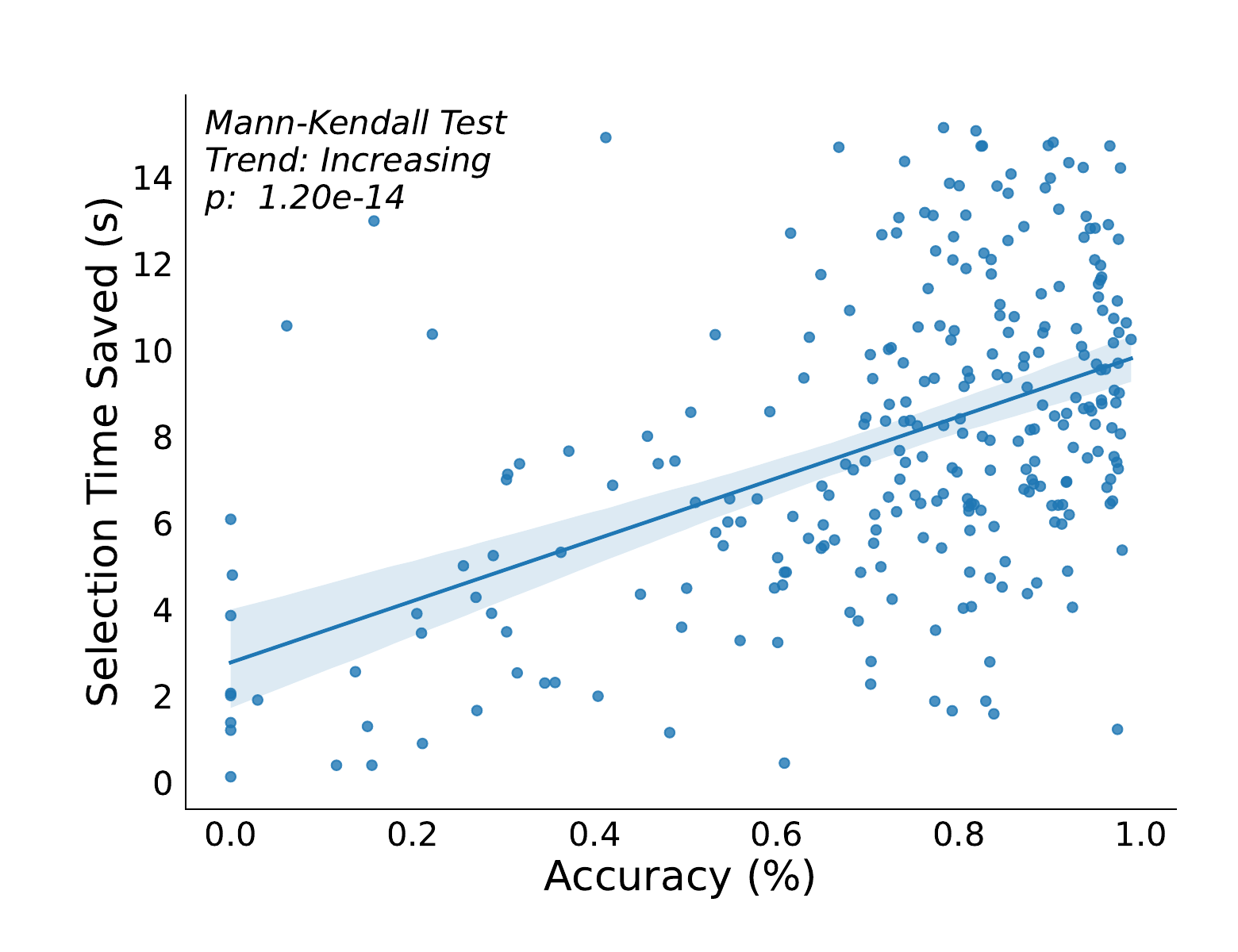}
  \caption{Average selection time saving versus accuracy for highest volume response templates.}
  \label{fig:ast_vs_acc}
\end{figure}

During the training phase of our model, we prioritize accuracy as an offline evaluation metric. However, in the online phase, we aim to optimize response selection time. Therefore, we seek a simple relationship between the two to inform model selection. To investigate this, we computed the accuracy of our model in our holdout that removes model predictions and compare it to the selection time savings between treatment and holdout groups for the (top $300$) response templates by volume. We observe a clear positive relationship validated by Mann-Kendall test such that as the prediction accuracy improves, the average time saved on selection tends to increase (Figure \ref{fig:ast_vs_acc}). Accurately predicting the most frequently occuring response templates is expected to significantly reduce our selection time. However, accurately predicting new response templates shortly after their introduction, as customer support advocates familiarize themselves with them, is also important for optimizing selection time.

\section{Conclusion}
In this work, we have described scaling laws for LLMs over closed domain adaptation with applications to a customer support use case. While most prior works on scaling laws has focused on the generative abilities of LLMs, we detail the effect on discriminative fine-tuning. Because existing deployed systems in industry are likely to involve a discrete classification, they can immediately benefit from the recent progress in LLMs including domain adaptation. Furthermore, this approach obviates safety concerns of LLMs in deployed systems such as hallucination by avoiding text generation altogether. In scenarios, where safety can be managed, such as those with a human in the loop to verify generated responses, practitioners can develop a domain-specific backbone. This backbone can be shared across across text classification and text generation deployments, or it can be gradually integrated into business operations as generative applications are adopted.

Using our experimental system, we show the benefit of more accurate models in our online case study, which is commonly a result of scaling up model size. However, for use cases with a human in the loop, larger models may introduce latency that negatively affects business operations. Given the relatively smooth linear scaling we observe with number of tokens in our domain adaptation experiments, smaller domain-specific LLMs are an attractive solution to increase accuracy without negatively affecting latency.

While our work has focused on customer support applications, we believe that our system is applicable in any situation where structured conversations or text classifications might occur. Furthermore, Our system would be applicable to open-domain settings with a fixed set of classification categories through the removal of the domain-adaptation step. Therefore, while the method was developed with the customer support use-case in mind, it is applicable to a wide variety of situations.

\bibliographystyle{ACM-Reference-Format}
\bibliography{../../shared/tex/paper}

\appendix

\section*{Appendices}

\subsection*{Dataset Formatting}
Here we show how transcripts from our customer support dataset are formatted. Figure~\ref{fig:DatasetFormatting} depicts how raw text is formatted for the pre-training and discriminative fine-tuning phases of our training pipeline.

\begin{figure}[h]
\raggedright
\newcommand{\bucketholder}[1]{\textless #1\textgreater}
\noindent\makebox[\columnwidth]{\rule{\columnwidth}{0.5pt}}\\
\noindent\textbf{Raw Text:} \\
\hspace*{1em}\textbf{Customer:} What are the balances on my accounts? \\
\hspace*{1em}\textbf{System:} Hi \bucketholder{NAME}, I'll get you to someone who can help. You don't have to wait. We'll notify you when they reply. \\
\hspace*{1em}\textbf{Customer:} Ty \\
\hspace*{1em}\textbf{Customer:} Just a general question... What are the totals of all my accounts with cash app? \\
\hspace*{1em}\textbf{...} \\
\hspace*{1em}\textbf{Advocate:} Is there anything else that I can do for you? \\
\hspace*{1em}\textbf{Customer:} No, that's it, Thanks! \\
\vspace{\baselineskip}

\noindent\textbf{Pre-training Sample:} \\
\bucketholder{CUSTOMER}: What are the balances on my accounts? \textbackslash n \bucketholder{SYSTEM}: Hi \bucketholder{NAME}, I'll get you to someone who can help. ... \textbackslash n \bucketholder{ADVOCATE}: Is there anything else that I can do for you? \textbackslash n \bucketholder{CUSTOMER}: No, that's it, Thanks! \\
\vspace{\baselineskip}

\noindent\textbf{Discriminative Fine-tuning Sample:} \\
\textbf{Input:} What are the balances on my accounts? Hi \bucketholder{NAME}, I'll get you to someone who can help. ... the totals of all my accounts with cash app? \\
\textbf{Classification Label:} \bucketholder{VIEW\_BALANCE\_TEMPLATE}

\noindent\makebox[\columnwidth]{\rule{\columnwidth}{0.5pt}}
\caption{Dataset Formatting for Pre-training and Discriminative Fine-tuning.}
\label{fig:DatasetFormatting}
\end{figure}

\begin{figure*}
	\centering     
	\subfigure[Domain Adaptation FLOPs vs  Top-5 Accuracy]{\label{fig:apendix-a}\includegraphics[width=57mm]{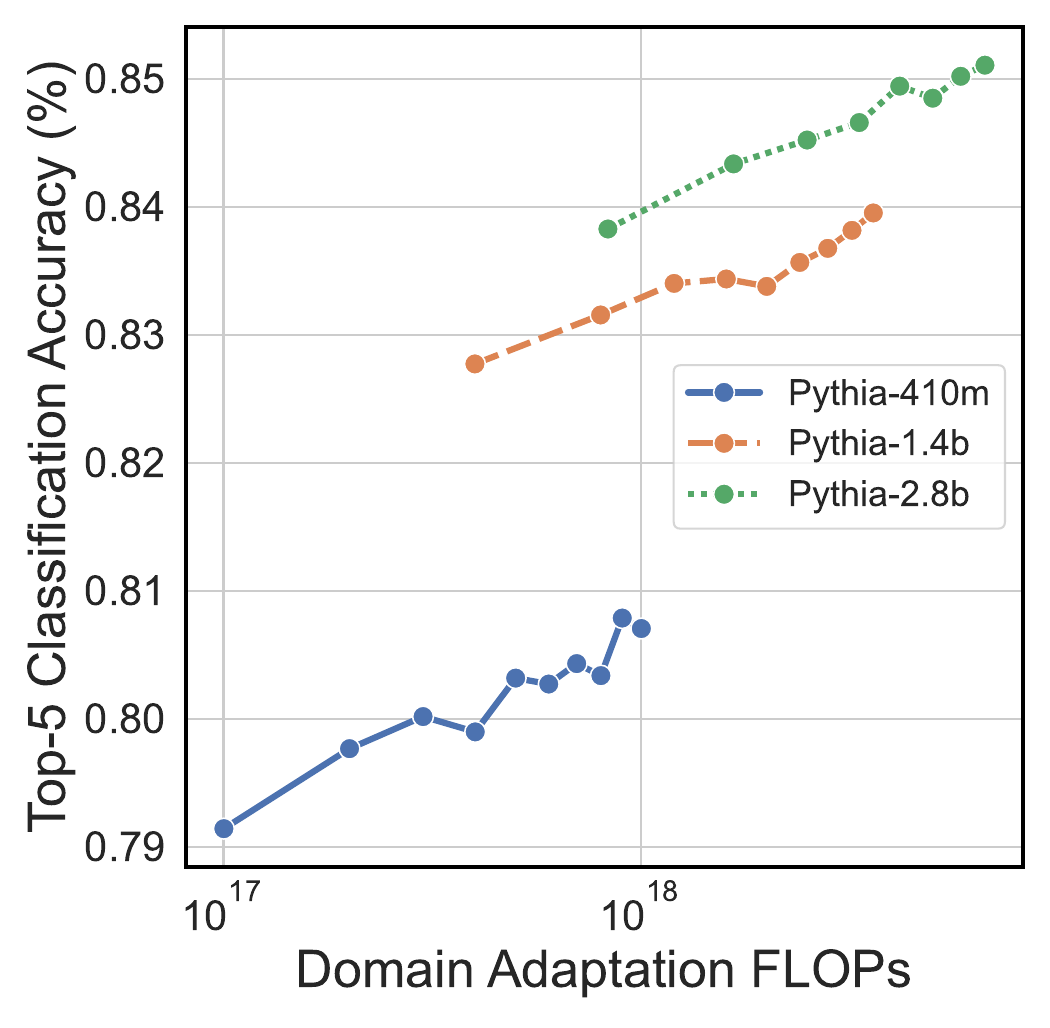}}
	\subfigure[Domain Adaptation Tokens vs  Top-5 Accuracy]{\label{fig:appendix-b}\includegraphics[width=56mm]{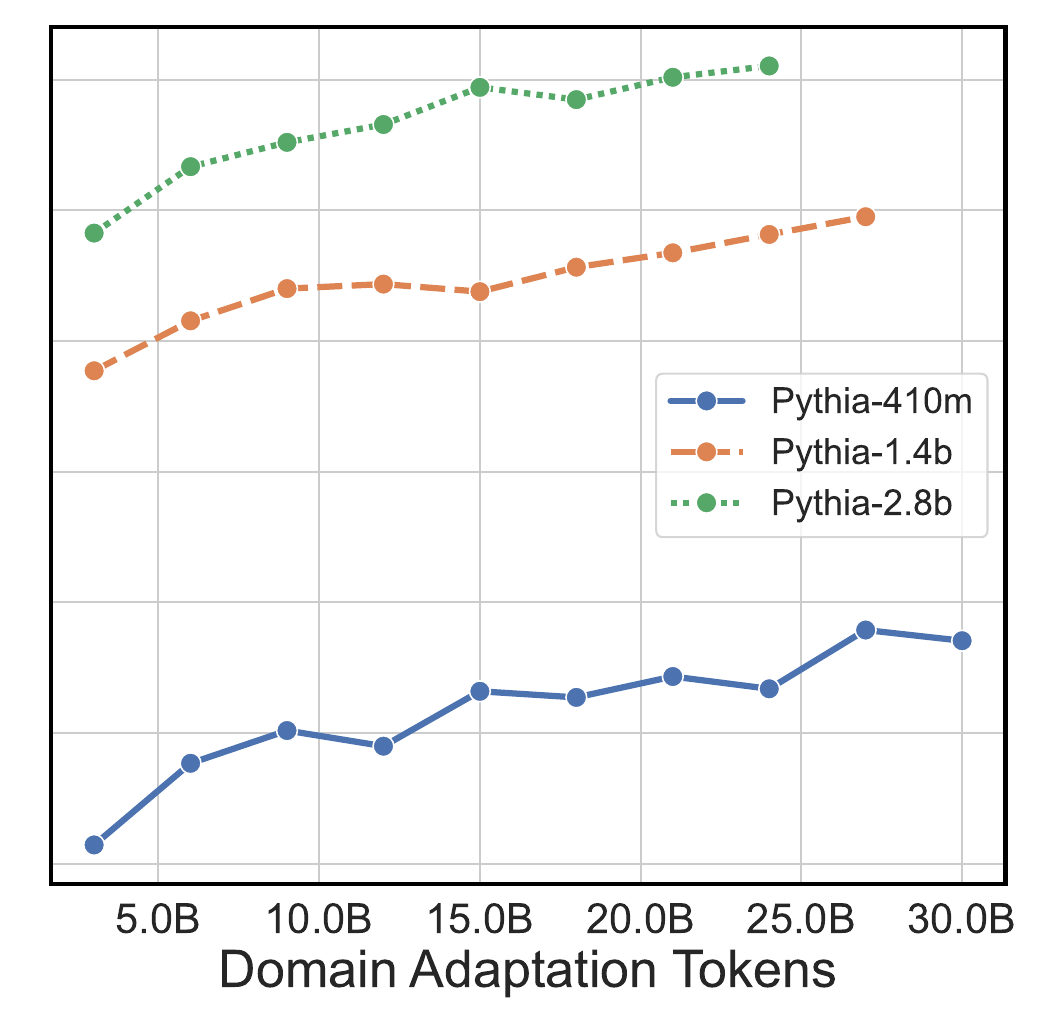}}
	\subfigure[Domain Adaptation Loss vs Top-5 Accuracy]{\label{fig:appendix-c}\includegraphics[width=56mm]{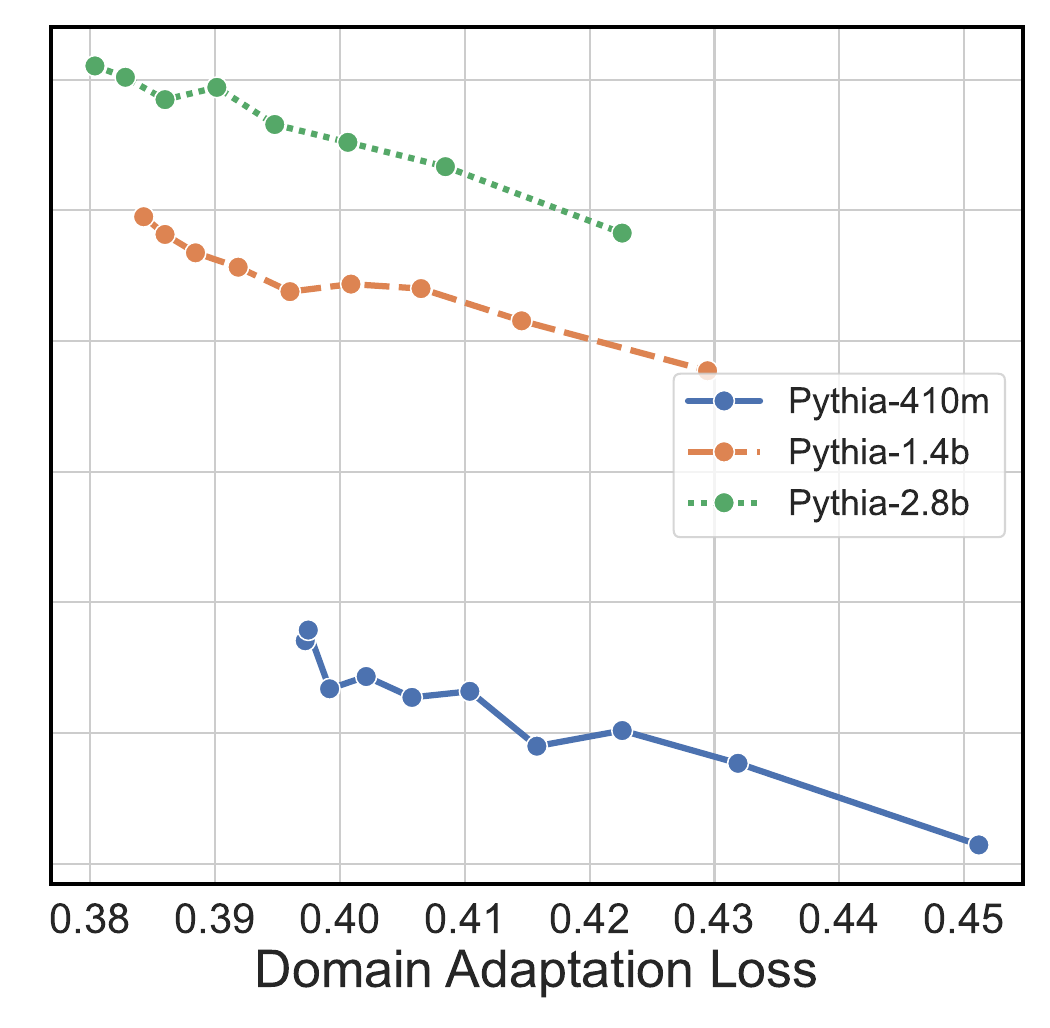}} 
	\caption{Discriminative fine-tuning empirical scaling properties across different model sizes. }
	\label{fig:empricalScaling_accuracy}
\end{figure*}

\subsection*{Domain Adaptation Scaling Results}
\begin{figure}
	\centering
	\subfigure[Domain adaptation training Loss on different steps]{\includegraphics[width=0.9\columnwidth]{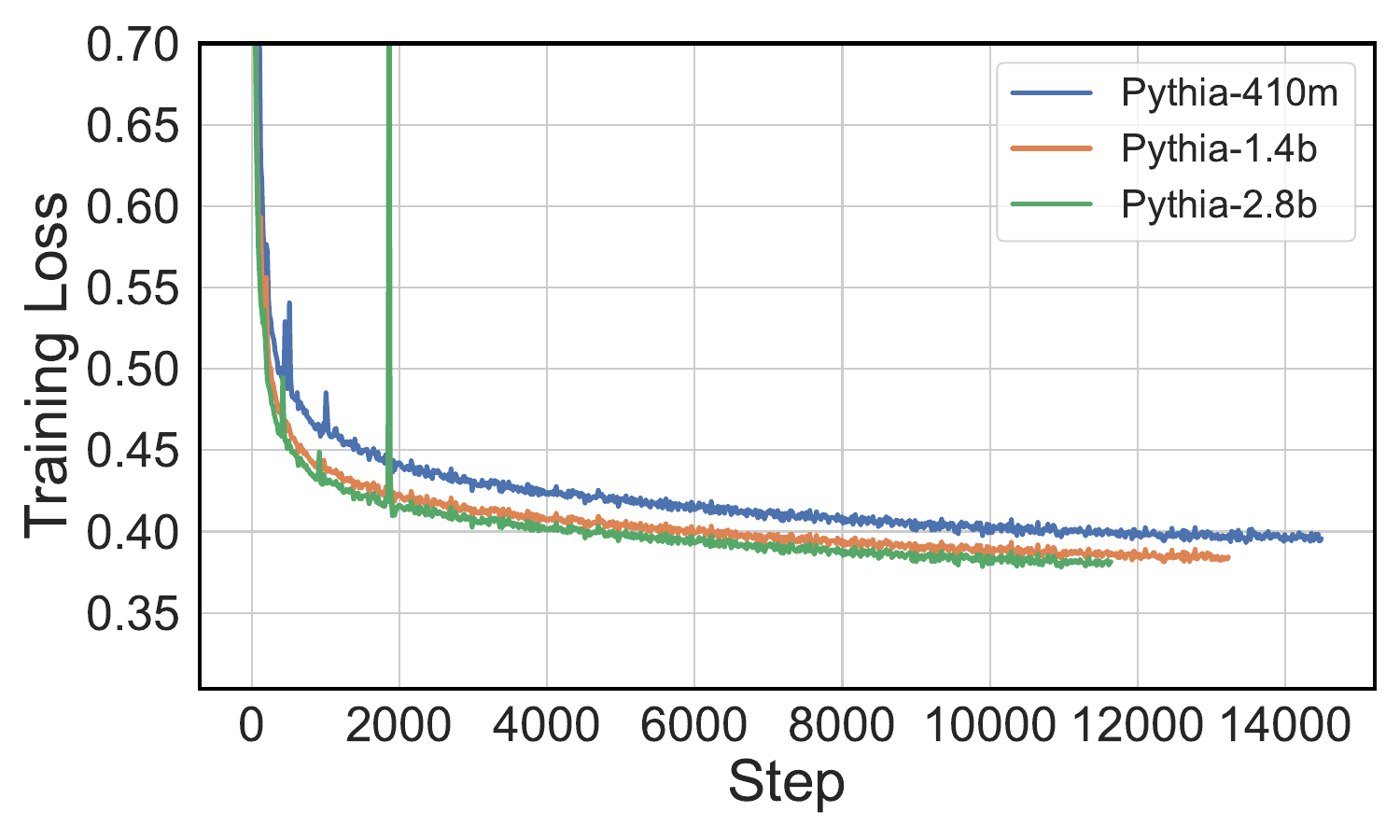}}
	\subfigure[Domain adaptation validation loss on different steps]{\includegraphics[width=0.9\columnwidth]{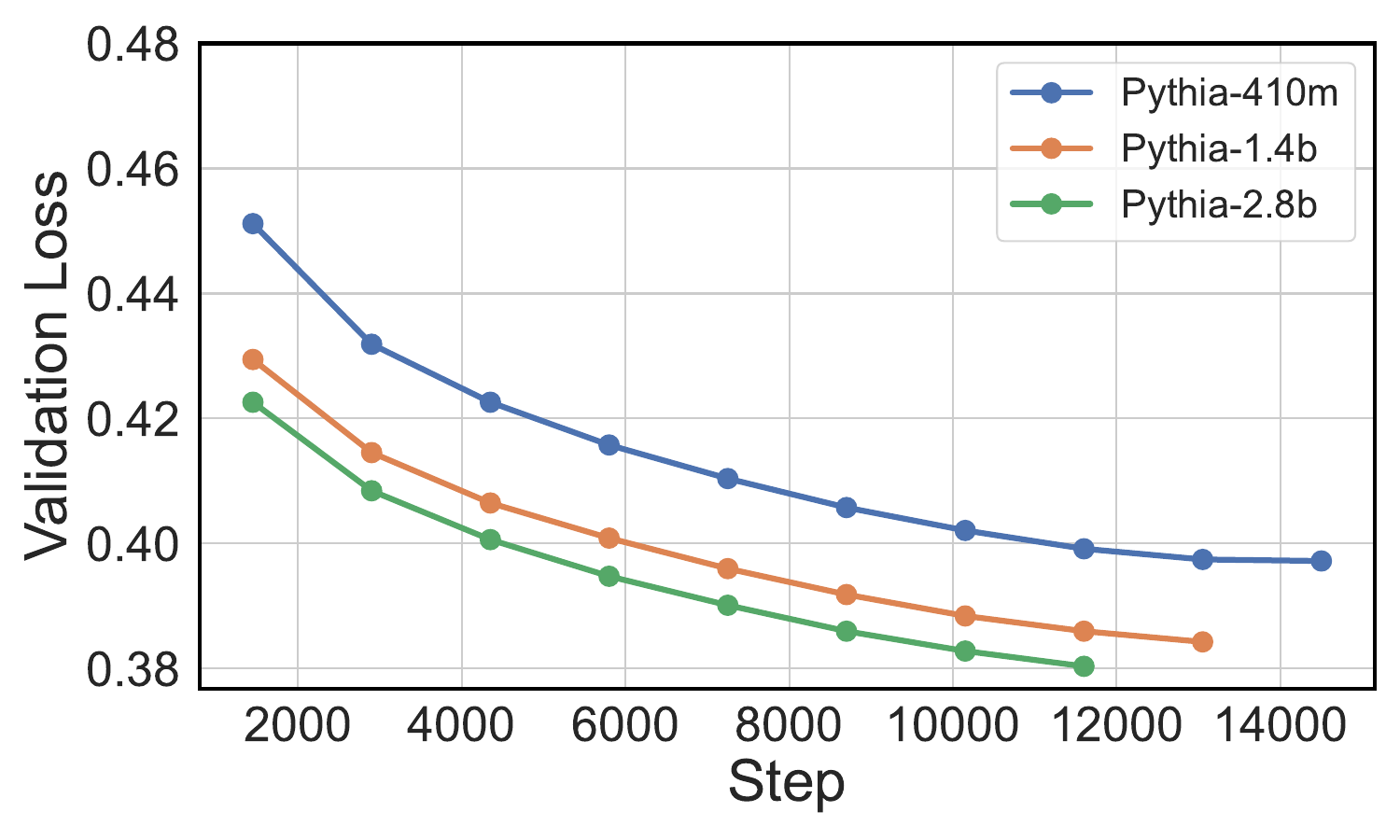}}
	\caption{Domain adaptation emprical scaling results across different model sizes. }
	\label{fig:DomainAdaptation}
\end{figure}

Here we look at the scaling properties of the domain adaptation step within our pipeline.  Figure~\ref{fig:DomainAdaptation} shows the training and validation loss of various models across different steps of pre-training.  

\FloatBarrier

\subsection*{Classification Accuracy in Scaling Results}
Figure~\ref{fig:empricalScaling_accuracy} depicts properties of the scaling laws observed in our experiments with regard to classification accuracy instead of loss in Figure~\ref{fig:empricalScaling}.

\end{document}